\documentclass[10pt,twocolumn,letterpaper]{article}

\usepackage{cvpr}              

\usepackage{textcomp}
\usepackage{multirow}
\pdfoutput=1

\DeclareUnicodeCharacter{2009}{\,}

\usepackage{enumitem}
\usepackage[most]{tcolorbox}
\tcbset{aibox/.style={
    breakable,
    break at=\maxdimen,
    height fixed for=first and middle,
}}
\newtcolorbox{AIbox}[2][]{aibox,title=#2,#1}
\newcommand{\DV}{Galaxy Walker\xspace}
\setlist[enumerate]{itemsep=0.5em, parsep=0pt, topsep=0.5em}

\definecolor{cvprblue}{rgb}{0.21,0.49,0.74}
\usepackage[pagebackref,breaklinks,colorlinks,allcolors=cvprblue]{hyperref}

\def\paperID{9614} 
\def\confName{CVPR}
\def\confYear{2025}

\title{Galaxy Walker: Geometry-aware VLMs For Galaxy-scale Understanding}

\author{
\textbf{Tianyu Chen$^{\star}$, Xingcheng Fu$^{\diamond}$, Yisen Gao$^{\dagger}$, Haodong Qian$^{\diamond}$, Yuecen Wei$^{\star \ddagger}$, Kun Yan$^{\star}$} \\ 
\textbf{Haoyi Zhou$^{\star \ddagger}$, Jianxin Li$^{\star}$} \\
SKLCCSE, School of Computer Science and Engineering, Beihang University, China$^{\star}$ \\
School of Software, Beihang University, China$^{\ddagger}$ \\
Key Lab of Education Blockchain and Intelligent Technology, Guangxi Normal University, China$^{\diamond}$\\
Institute of Artificial Intelligence, Beihang University, Beijing, China$^{\dagger}$
\\
{\tt\small \{tianyuc, zhouhy, lijx\}@buaa.edu.cn}
}

\begin{document}
\maketitle
\footnotetext[1]{The corresponding author is Jianxin Li (lijx@buaa.edu.cn).}
\begin{abstract}
Modern vision-language models (VLMs) develop patch embedding and convolution backbone within vector space, especially Euclidean ones, at the very founding. When expanding VLMs to a galaxy scale for understanding astronomical phenomena, the integration of spherical space for planetary orbits and hyperbolic spaces for black holes raises two formidable challenges. a) The current pre-training model is confined to Euclidean space rather than a comprehensive geometric embedding. b) The predominant architecture lacks suitable backbones for anisotropic physical geometries. In this paper, we introduced Galaxy-Walker, a geometry-aware VLM, for the universe-level vision understanding tasks. We proposed the geometry prompt that generates geometry tokens by random walks across diverse spaces on a multi-scale physical graph, along with a geometry adapter that compresses and reshapes the space anisotropy in a mixture-of-experts manner. Extensive experiments demonstrate the effectiveness of our approach, with Galaxy-Walker achieving state-of-the-art performance in both galaxy property estimation ($R^2$ scores up to $0.91$) and morphology classification tasks (up to $+0.17$ F1 improvement in challenging features), significantly outperforming both domain-specific models and general-purpose VLMs.

\end{abstract}    
\section{Introduction}







\begin{figure}[t] 
    \centering
    \includegraphics[width=0.48\textwidth]{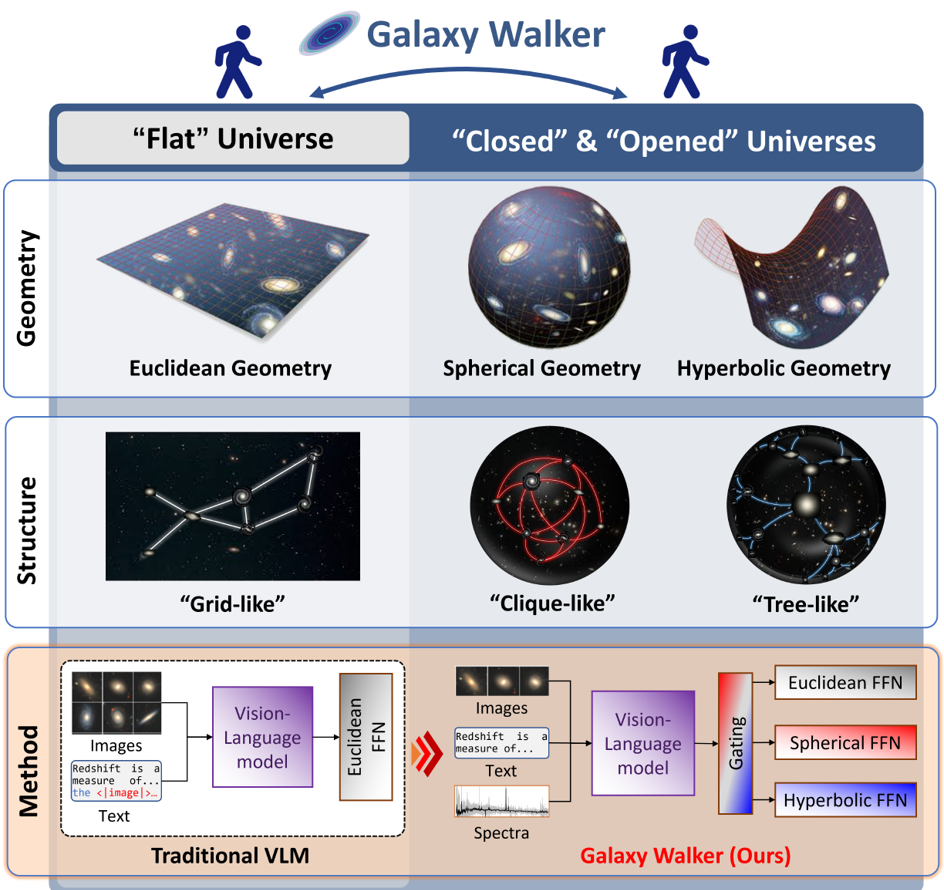}
    \caption{\textbf{Geometries of the universe.} While traditional VLMs are confined to flat Euclidean space, the actual universe exhibits rich geometric diversity including spherical and hyperbolic spaces, motivating our Galaxy Walker framework to incorporate multi-geometric representations.}
    \label{fig:intro}
\end{figure}



Geometric cognition has always been an essential problem in ordinary imaging, like overlap~\cite{overlap}, scaling~\cite{scaling}, and perspective~\cite{perspective}. It becomes a more challenging one in Astronomy Imagery, where the distance between pixels no longer stands for its ``real'' distance, and the measurement scales up to the galaxy-level. Backing to the Eddington experiment in 1919, we still view the Solar system as a flat Euclidean space~\cite{flatuniverse} in telescope and measure the Gravity Lensing phenomenon, while the LIGO project~\cite{LIGO} detects the gravitational waves of astronomical events with hyperbolic space starting from $2002$. One key difference is that the physicists~\cite{universegeo1,universegeo2} have gradually unveiled the rich geometric diversity at the galaxy scale. As illustrated in Figure~\ref{fig:intro}, if we aim to walk from the flat universe to the other universes, the geometric cognition must extend from Euclidean space to diverse ones for better lare-scale understanding.


To take a concrete investigation on this problem, we apply the predominant Vision-Language Models (VLMs), e.g. GPT4-o~\cite{openai2024gpt4technicalreport} and Claude3.5~\cite{claude} on astronomical analysis tasks. The complete results are included in Table~\ref{tab:model_comparison}, they achieve $R^2$ scores below $0.6$ in galaxy property estimation tasks, while their F1 scores in morphology classification hover between $0.4$-$0.7$. The performance degradation mainly comes from two constraints in VLM architectures: $(1)$ At the foundational level, patch embedding and convolutional backbones are constructed within Euclidean vector spaces, struggling to effectively represent non-Euclidean geometric features; $(2)$ At the feature fusion level, self-attention and FFN mechanisms tend to model token relationships based on planar distances, overlooking geometric measures like spherical and hyperbolic distances.

\begin{figure*}[!t]
    \centering
    \includegraphics[width=\linewidth]{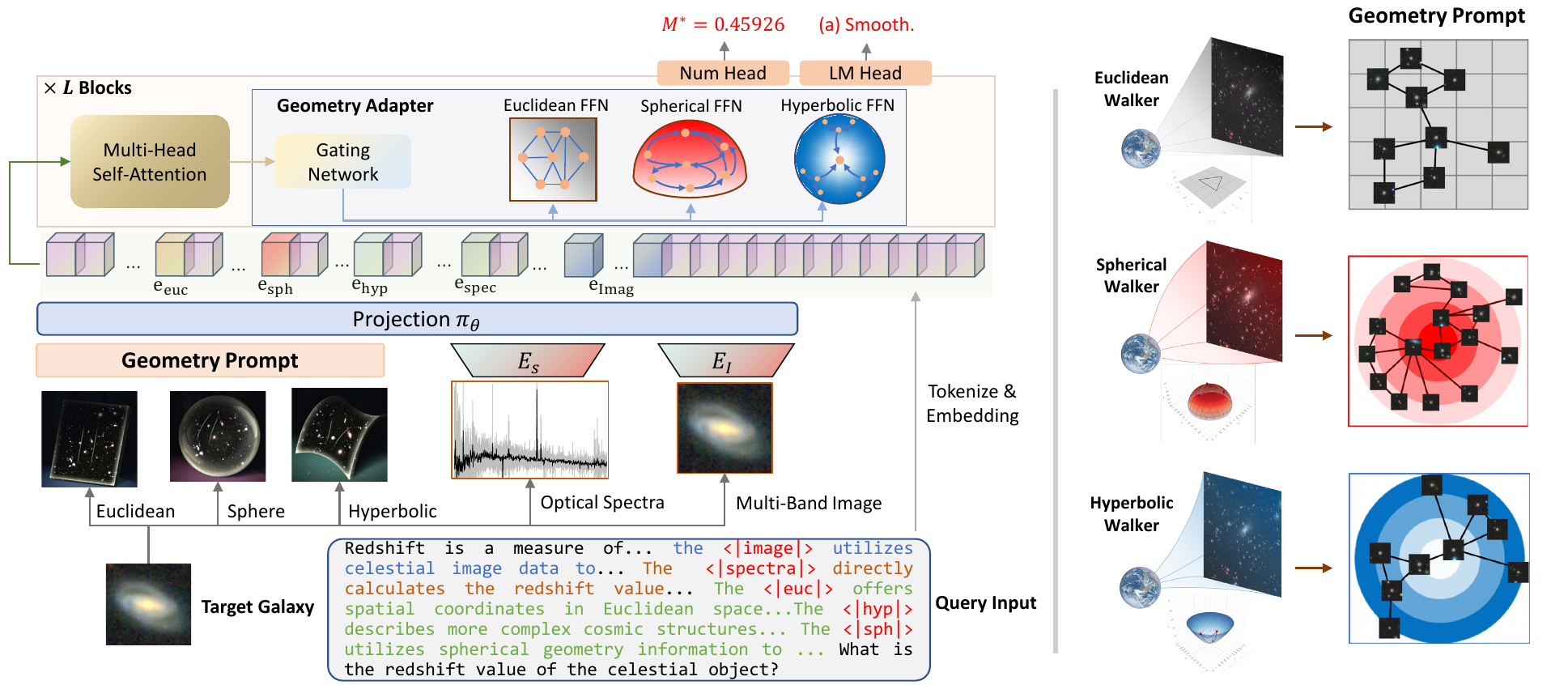}
    \caption{\textbf{The overall framework of Galaxy Walker. }\textbf{Left:} The architecture integrates a Geometry Adapter with the pre-trained VLM backbone. The adapter includes a projection layer $\pi_\theta$ that processes various input modalities (e.g., Euclidean, Spherical, Hyperbolic embeddings, spectral data, and multi-band images), followed by $L$ transformer blocks enhanced with geometry-aware FFN experts. A gating network dynamically routes features to appropriate geometric experts. Two parallel heads (Numeric Head and LM Head) enable both regression and classification tasks. \textbf{Right:} Visualization of how different geometric spaces (Euclidean, Spherical, and Hyperbolic Walker) process astronomical data, demonstrating the distinct token arrangements and relationships in each geometry. The Geometry Prompt guides the model to utilize appropriate geometric representations for different astronomical features.} 
    \label{fig:framework}
\end{figure*}

To address these challenges, we present Galaxy-Walker, the first geometry-aware VLM framework designed for galaxy-scale understanding. As illustrated in Figure \ref{fig:framework}, Galaxy-Walker introduces two key innovations: $(1)$ A Geometry Prompt mechanism that generates geometric tokens through random walks on multi-scale physical graphs across Euclidean, spherical, and hyperbolic spaces, injecting diverse geometric priors at the input level; $(2)$ A Geometry Adapter incorporating Euclidean, spherical, and hyperbolic FFN expert modules, which adaptively processes different geometric features through a mixture-of-experts approach, effectively reshaping spatial anisotropy.

Extensive experiments validate the effectiveness of Galaxy-Walker. In galaxy property estimation tasks, our method achieves $R^2$ scores ranging from $0.52$ to $0.91$, surpassing general VLMs by $50$-$80$ percentage points and significantly outperforming domain-specific models like AstroCLIP. In morphology classification, Galaxy-Walker demonstrates substantial improvements ($+0.17$) in recognizing characteristic features such as BAR and SAC, showcasing its robust geometric understanding capabilities. These results indicate that introducing geometry awareness into VLM architectures is crucial for advancing galaxy-scale understanding tasks.

Our main contributions can be summarized as follows: $(1)$ We identify and propose a solution to address the geometric representation limitations in VLMs when processing astronomical data; $(2)$ We design a geometry-aware framework with novel prompt and adapter mechanisms to capture diverse geometric features in galaxy observations effectively; $(3)$ We demonstrate empirical improvements across multiple astronomical tasks, suggesting a promising direction for enhancing VLMs' capability in domain-specific applications.

\section{Related Work}

\textbf{Astronomical machine learning} has evolved from early supervised learning applications~\cite{galaxysurvey} to more sophisticated approaches, including unsupervised methods for gravitational lens detection and anomaly identification~\cite{Melchior2022AutoencodingGS,stein2021selfsupervisedsimilaritysearchlarge}. Recent advances integrate convolutional networks for image analysis and MLPs for luminosity data processing~\cite{Pasquet2018PhotometricRF}, with AstroCLIP pioneering cross-modal galaxy feature interaction modeling~\cite{Astroclip}.

\textbf{Vision Language Models} have revolutionized multimodal understanding through joint visual-textual representations~\cite{zhang2024visionlanguagemodelsvisiontasks, PaLM-E}. CLIP~\cite{Clip} demonstrated exceptional performance through self-supervised training on $400$M image-text pairs, while models like GPT4-o~\cite{openai2024gpt4technicalreport}, Claude 3.5~\cite{claude}, and LLAMA-3.2-VL~\cite{llama} continue advancing vision-language integration capabilities.  Recent works like GeoCode and GeoGPT4V have explored enhancing VLMs' geometric perception through data augmentation, aiming to address fundamental geometry-related question-answering tasks~\cite{geocode, geogpt4v}.

\textbf{Geometric deep learning} leverages Riemannian geometry for complex data modeling. Hyperbolic geometry has proven effective for hierarchical image relationships~\cite{hypervision,khrulkov2020hyperbolicimageembeddings}, while 
the spherical space is used to capture the global view~\cite{sphereresolution,  sphere2, sphere3}. Mixed curvature spaces~\cite{mixturecurve} allow for flexible learning representations across a range of Riemannian manifolds. Mapping image features onto Riemannian manifolds~\cite{Wang2022DreamNetAD} enhances the classification and comprehension of complex images.

\section{Method}
Galaxy Walker is a geometry-aware framework that enhances pre-trained VLMs with non-Euclidean geometric priors for comprehensive astronomical understanding. As illustrated in Figure \ref{fig:framework}, our framework consists of two key components: a \textbf{Geometry Prompt} that generates geometry tokens by random walks across diverse spaces on a multi-scale physical graph, along with a \textbf{Geometry Adapter} that orchestrates multi-modal representations through a mixture-of-experts architecture spanning euclidean, spherical, and hyperbolic spaces. To facilitate numerical predictions while preserving the model's language modeling capabilities, we augment the architecture with a dedicated Numeric Head alongside the original LM (Language Modeling) Head, enabling both regression and classification tasks in astronomical analysis.

\subsection{Geometry Prompt}
The geometric properties of universe space are crucial for understanding the large-scale structure of the universe~\cite{universegeo2}.
It can provide large models with insights into the diverse and complex underlying structures between galaxies. 
By leveraging these geometric insights, we design a geometry prompt mechanism to guide the model to leverage appropriate geometric spaces for different astronomical features through specialized geometric walkers. 
Each walker (Euclidean, Spherical, and Hyperbolic) processes astronomical data in its corresponding geometric space, capturing distinct spatial relationships and structural patterns that complement the Geometry Adapter's expert layers.

\textbf{Multi-View Structure Construction of Galaxies.} 
The vast expanse of universe space can be understood as a complex Riemannian geometric space~\cite{universegeo2}, which can encompass different geometries, including Euclidean ($\mathbb{E}$), Hyperbolic ($\mathbb{H}$), and Spherical ($\mathbb{S}$) spaces.  A Euclidean space, characterized by zero curvature, represents an infinitely flat universe.  In this scenario, galaxies are highly influenced by nearby local structures.  In contrast, a hyperbolic space, with its negative curvature, captures the universe's accelerated expansion, providing a natural framework to trace the evolution of galaxies and uncover hierarchical relationships within their development.  Meanwhile, spherical space, defined by positive curvature, suggests a universe with a closed, global topology, enhancing our ability to capture the overall similarity among galaxies at a global scale.  To gain a deeper and more comprehensive understanding of the universe's structure and behavior, we will explore three distinct curvatures of universe space and build models to capture the relationships among galaxies within each geometry.

As shown in Fig.~\ref{fig:framework} right, we start by constructing a multi-relational graph of galaxies based on their physical positions $\mathbf{V}_{phy}$, which consist of Right Ascension (RA) and Declination (DEC). First, we establish universe coordinates from different geometric perspectives to derive relationships between galaxies. Specifically, a universe geometric coordinate $\mathbf{V}_\mathbb{M}$ is first mapped to the tangent space $\mathbb{T}\mathbb{M}$ at the origin by the projection function $proj$, and then it is transformed to the manifold $\mathbb{M}$ ($\mathbb{M} \in \{\mathbb{E}, \mathbb{H}, \mathbb{S} \}$) by the exponential mapping $exp_o^c$:
\begin{equation} \mathbf{V}_{\mathbb{M}} = exp_o^c(proj(\mathbf{V}_{phy})), \end{equation}
where $c$ represents the curvature of the manifold $\mathbb{M}$.
Subsequently, we identify galaxies with similar positions using K-Nearest Neighbors (KNN)~\cite{knn} based on these universe coordinates, forming the relational graphs $\mathbb{A}_\mathbb{M}$.

\textbf{Geometry-aware Feature Learning.} After getting a graph $G$ with node feature matrix $\mathbf{X} \in \mathbb{R}^{n \times d}$ and adjacency matrix $\mathbf{A}_\mathbb{M}$, we study their representation in three different geometric universes. Since the node feature $\mathbf{X}$ (like image/spectrum) is in Euclidean space, we first need to translate it to the corresponding manifold $\mathbb{M}$ (Step1). The geometric prompt is then obtained using the relational graph $\mathbb{A}_\mathbb{M}$ in the manifold (Step2). 
The geometry-aware features $\mathbf{P}_{\mathbb{M}}$ are learned through a two-layer architecture:
\begin{equation}
 \left\{\begin{matrix} 
\mathbf{Z}_{\mathbb{M}}= \mathcal{F}_{\mathbb{E}\to \mathbb{M}}(\mathbf{X},\mathbf{A}_{\mathbb{M}}), \quad \text{Step1}\\
\mathbf{P}_{\mathbb{M}}= \mathcal{F}_{\mathbb{M} \to \mathbb{M}}(\mathbf{Z}_{\mathbb{M}},\mathbf{A}_{\mathbb{M}}), \quad \text{Step2}
\end{matrix}\right.    
\end{equation}
where $\mathcal{F}_{\mathbb{M}_\text{in} \to \mathbb{M}_\text{out}}$ represents a Riemannian GraphSAGE layer defined as:
\begin{equation}
\mathcal{F}_{\mathbb{M}_\text{in} \to \mathbb{M}_\text{out}}(\mathbf{X},\mathbf{A}_{\mathbb{M}_\text{in}}) = \exp_o^{c_{out}}(\text{SAGE}(\log_o^{c_{in}}(\mathbf{X}), \mathbf{A}_{\mathbb{M}_\text{in}}),
\end{equation}
where $log_o^c$ denotes logarithmic mapping, which maps the feature $\mathbf{X}$ located on the manifold $\mathbb{M_\text{in}}$ to the tangent plane of the origin. Sage denotes the GraphSage network\cite{Sage}. The specific formulation forms related to Riemannian manifold operators are reported in the Appendix.


\subsection{Geometry Adapter}

The Geometry Adapter is designed to handle spatial anisotropy through a mixture-of-experts (MoE) approach, incorporating domain-specific geometric priors from astronomical observations, while preserving the conventional Euclidean FFN from pre-trained VLMs as one expert:
\begin{equation}
    \mathcal{F}_E(x) = W_2(\sigma(W_1x + b_1)) + b_2.
\end{equation}

We then introduce specialized experts for spherical and hyperbolic geometries to enhance the model's capability in processing astronomical spatial relationships. Inspired by metric theories in non-Euclidean spaces \cite{bronstein2017geometric}, we design spherical and hyperbolic expert layers that form a heterogeneous MoE network. The spherical expert layer is formulated as:
\begin{equation}
    \mathcal{F}_S(x) = \kappa \cdot \text{normalize}(W_2(\sigma(W_1x + b_1)) + b_2),
\end{equation}
where $\kappa$ controls the curvature and normalize $(\cdot )$ ensures the output lies on the unit sphere, effectively capturing angular relationships in celestial observations. The hyperbolic expert layer is defined as:
\begin{equation}
    \mathcal{F}_H(x) = \exp_0(W_2(\sigma(\log_0(W_1 \otimes x + b_1))) + b_2),
\end{equation}
where $\exp_0$ and $\log_0$ are the exponential and logarithmic maps at the origin of the Poincaré ball, enabling the model to process hierarchical astronomical structures.

For routing multimodal tokens to appropriate experts, we introduce a learnable Gating Network $G$ that computes routing probabilities:
\begin{equation}
    y = \sum_{i \in \{E,S,H\}} G_i(x) \cdot \mathcal{F}_i(x),
\end{equation}
where $G_i(x)$ represents the routing weight for expert $i$, and $\sum_i G_i(x) = 1$. The Geometry Adapter is inserted in every $k$ layer (e.g., $k = 4$) in the pre-trained VLM to maintain computational efficiency while providing sufficient geometric modeling capacity.

To facilitate geometric feature input, we initialize modality-specific projection layers $\pi_\theta$ that align different feature types with the token embedding space. For regression tasks, we augment the model with a learnable Numeric Head that performs numerical predictions through the last token's logits. The original LM Head is retained for classification tasks, maintaining the model's versatility across different astronomical applications.

\subsection{Two-stage Training}
We adopt a two-stage training strategy to optimize our geometry-aware framework effectively.

\textbf{Stage I: Geometric Prompt Learning.} The Geometry Prompt module is first trained independently using galaxy property estimation tasks (e.g., redshift prediction) to learn the geometric representations across three spaces. This stage ensures the model captures essential geometric patterns.

\textbf{Stage II: Geometry Adapter Learning.} We then train the Geometry adapter module while freezing all attention blocks for efficiency. Let $\mathcal{T}_r$ and $\mathcal{T}_c$ denote the sets of regression and classification tasks, respectively. For each task $t \in {\mathcal{T}_r \cup \mathcal{T}_c}$, we construct task-specific prompts using special modality tokens (details in Appendix). The training objective combines:
\begin{equation}
\mathcal{L} = \mathcal{L}_{LM} + \lambda\mathcal{L}_{reg},
\end{equation}

\noindent where $\mathcal{L}_{LM}$ is the language modeling loss and $\mathcal{L}_{reg}$ is the smooth L1 loss for regression tasks, computed between the numerical head predictions and ground truth values. $\lambda$ balances the two objectives.

To enhance numerical stability and modal representation, modal inputs are processed in fp32 precision with L2 normalization, followed by learnable scaling factors $\alpha_m$ applied after projection: $\mathbf{e}_m = \alpha_m\pi_m(\mathbf{x}_m)$. We further incorporate a replay mechanism that accumulates projected embeddings to the last token hidden states: $\mathbf{h}_{last} = \mathbf{h}_{last} + \sum_m\mathbf{h}_m$. The trainable parameters in this stage are constrained to modal-specific projection layers ${\pi_m}$,  Geometry Adapter FFN layers, numerical head for regression tasks, and scaling factors ${\alpha_m}$.

\begin{table*}[htbp]
\centering
\caption{\textbf{Statistics of our dataset for property estimation and morphological classification tasks.} The rightmost column shows the total number of classification samples for each split.}
\label{tab:datasets}
\resizebox{\textwidth}{!}{%
\begin{tabular}{l|r|rrrrrrrrrr|r}
\toprule
\multirow{2}{*}{\textbf{Split}} & \multirow{2}{*}{\textbf{\begin{tabular}[c]{@{}c@{}}Property\\ Estimation\end{tabular}}} & \multicolumn{11}{c}{\textbf{Morphological Classification}} \\
\cmidrule(l){3-13}
 & & \textbf{SMH} & \textbf{DEO} & \textbf{SPR} & \textbf{BAR} & \textbf{BLG} & \textbf{RND} & \textbf{EOB} & \textbf{SWP} & \textbf{SAC} & \textbf{MRG} & \textbf{Subtotal} \\ 
\midrule
Train & 84,121 & 389 & 29,538 & 31,276 & 31,276 & 31,276 & 16,207 & 25,246 & 18,745 & 18,745 & 129 & 202,827 \\
Evaluate & 21,051 & 98 & 9,587 & 9,541 & 9,541 & 9,541 & 9,332 & 6,707 & 5,942 & 5,942 & 2,508 & 68,739 \\
\bottomrule
\end{tabular}%
}
\end{table*}

\begin{table}[htbp]
    \centering
    \caption{\textbf{Statistics of our multi-relational graph constructed in different geometric spaces.} Each space maintains consistent node count and feature dimensionality while varying in edge connectivity to capture distinct geometric relationships.}
    \label{tab:graph}
    \begin{tabular}{l|rrr}
        \toprule
        \textbf{Geometry} & \textbf{\#Nodes} & \textbf{\#Edges} & \textbf{\#Features} \\
        \midrule
        Euclidean  & 105,172 & 720,665 & 1,024 \\
        Hyperbolic & 105,172 & 699,553 & 1,024 \\
        Spherical  & 105,172 & 722,081 & 1,024 \\
        \bottomrule
    \end{tabular}
\end{table}

\section{Experiment}

\begin{table*}[htbp]
    \centering
    \renewcommand{\arraystretch}{1.0}
    \resizebox{\textwidth}{!}{
        \begin{tabular}{l|cccc|cccccccccc}
            \toprule
            \multirow{2}{*}{\textbf{Models}} & \multicolumn{4}{c|}{\textbf{Property Estimation ($R^2$ Score)}} & \multicolumn{10}{c}{\textbf{Morphology Classification (F1 Score)}} \\
            \cmidrule(lr){2-5} \cmidrule(lr){6-15}
            & $\mathbf{M_*}$ & $\mathbf{Z_{MW}}$ & $\mathbf{t_{age}}$ & \textbf{sSFR} & \textbf{SMH} & \textbf{DEO} & \textbf{SPR} & \textbf{BAR} & \textbf{BLG} & \textbf{RND} & \textbf{EOB} & \textbf{SWP} & \textbf{SAC} & \textbf{MRG} \\
            \midrule
            \multicolumn{15}{l}{\textit{Baselines-Domain Specific Methods}} \\
            Photometry(MLP) & 0.67 & 0.41 & 0.27 & 0.34 & -- & -- & -- & -- & -- & -- & -- & -- & -- & -- \\
            Stein & -- & -- & -- & -- & 0.68 & 0.81 & 0.95 & 0.37 & 0.77 & \underline{0.81} & 0.75 & \underline{0.76} & 0.44 & 0.71 \\
            AstroCLIP(Zero-shot) & 0.87 & 0.57 & 0.43 & 0.63 & -- & -- & -- & -- & -- & -- & -- & -- & -- & -- \\
            AstroCLIP(Fine-tuned) & \underline{0.88} & \underline{0.64} & \underline{0.47} & \underline{0.69} & \textbf{0.83} & \underline{0.97} & \underline{0.96} & \underline{0.54} & \underline{0.81} & 0.79 & \underline{0.84} & 0.73 & \underline{0.47} & \underline{0.73} \\
            \midrule
            \multicolumn{15}{l}{\textit{Baselines-General VLMs}} \\
            GPT4-o & 0.05& 0.43& -0.48& 0.59& 0.38& 0.67& 0.63& 0.42& 0.37& 0.49& 0.62& 0.57& 0.38& 0.47\\
            Claude-3.5 & -0.03& 0.20& -0.61& 0.62& 0.34& 0.33& 0.24& 0.42& 0.41& 0.40& 0.66& 0.39& 0.37& 0.59\\
            Llama-3.2-90B-Vision-Instruct & -2.99& -5.46& 0.28& 0.34& 0.38& 0.42& 0.65& 0.37& 0.42& 0.49& 0.66& 0.58& 0.46& 0.60\\
            \midrule
            \multicolumn{15}{l}{\textit{Our Method - Geometry-aware VLM}} \\
            Galaxy Walker & \textbf{0.91} & \textbf{0.69} & \textbf{0.52} & \textbf{0.84} & \underline{0.76} & \textbf{0.97} & \textbf{0.96} & \textbf{0.71} & \textbf{0.83} & \textbf{0.82} & \textbf{0.87} & \textbf{0.79} & \textbf{0.64} & \textbf{0.77} \\
            Improvements & +0.03 & +0.05 & +0.05 & +0.15 & -0.07 & 0.00 & 0.00 & +0.17 & +0.02 & +0.01 & +0.03 & +0.06 & +0.17 & +0.04 \\
            \bottomrule
        \end{tabular}
    }
    \caption{\textbf{Comprehensive evaluation of different models on galaxy property estimation ($R^2$ score) and morphology classification (F1 score) tasks.} Our proposed Galaxy Walker demonstrates superior performance across most metrics, achieving state-of-the-art results in all property estimation tasks and several morphology classification categories. The previous best results are \underline{underlined}, while our \textbf{best} results are shown in bold. Notably, Galaxy Walker shows significant improvements in sSFR estimation ($+0.15$) and morphological features like BAR and SAC classifications ($+0.17$), while maintaining competitive performance in other categories.}
    \label{tab:model_comparison}
\end{table*}

\subsection{Experiment Settings}

\textbf{Tasks and Datasets.} Our evaluation framework encompasses two primary astronomical analysis tasks: Property Estimation and Morphology Classification~\cite{Astroclip}. Property Estimation targets four fundamental galaxy attributes ($\mathbf{M^*}$, $\mathbf{Z_{MW}}$, $\mathbf{t_{age}}$, $\mathbf{sSFR}$), essential for understanding galaxy evolution and stellar populations.~\cite{galaxysurvey} Morphology Classification examines ten distinct structural features (SMH, DEO, SPR, BAR, BLG, RND, EOB, SWP, SAC, MRG), enabling a comprehensive assessment of galactic structures across scales and orientations~\cite{galaxy_zoo}. Following AstroCLIP's methodology~\cite{Astroclip}, we structure these tasks as question-answering problems to facilitate unified VLM training and inference. As shown in Table \ref{tab:datasets}, Our dataset consists of $84,121$ training samples and $21,051$ evaluation samples for property estimation, and a total of $271,566$ samples for morphological classification, with varying sample distributions across morphological classes.

\textbf{Input Modalities and Task Formulation.}
Property Estimation tasks integrate multi-band images from DESI-LS DR9~\cite{Hahn_2023a}, spectra from DESI EDR Survey~\cite{spectrum}, and geometric features from three spaces, outputting continuous property predictions. Morphology Classification, constrained by Galaxy Zoo DECaLS5 ~\cite{galaxy_zoo} dataset limitations, utilizes only multi-band images and geometric features for categorical predictions.

\textbf{Multi-geometric Graph Construction.}
We construct three complementary graphs based on galaxies' physical coordinates (Right Ascension and Declination) through k-nearest Neighbor connections. As detailed in Table \ref{tab:graph}, these graphs span Euclidean (720,665 edges), Hyperbolic ($699,553$ edges), and Spherical spaces ($722,081$ edges), each containing $105,172$ nodes with $1,024$-dimensional features derived from matched DESI-LS galaxy images and DESI spectra pairs. This multi-geometric approach enables comprehensive capture of diverse spatial relationships in astronomical structures.

\textbf{Other Implementation Details.} We initialize the VLM of GalaxyWalker from Qwen2-VL-2B-Instruct~\cite{Qwen}. We design training and inference templates for each task and the details are included in supplementaries.

\textbf{Baselines.}
We evaluate our approach against both domain-specific models and general-purpose VLMs. For domain-specific baselines, we include: Photometry~\cite{Pasquet2018PhotometricRF}, a classical MLP-based approach representing traditional machine learning methods for astronomical property estimation; Stein~\cite{stein2021selfsupervisedsimilaritysearchlarge}, a specialized self-supervised model optimized for scientific data analysis; and AstroCLIP~\cite{Astroclip}, a pioneering astronomical multimodal model evaluated in both zero-shot and fine-tuned settings. For general-purpose VLMs, we compare against state-of-the-art closed-source models (GPT4-o~\cite{openai2024gpt4technicalreport} and Claude-3.5~\cite{claude}) and the leading open-source multimodal model (Llama-3.2 ~\cite{llama}), representing the current frontier of large-scale vision-language understanding capabilities.

\subsection{Main Results}

As depicted in Table \ref{tab:model_comparison}, our extensive experimental evaluation demonstrates the superior performance of Galaxy Walker across both galaxy property estimation and morphology classification tasks, validating our geometry-aware approach to galaxy-scale understanding.

\textbf{Limitations of General VLMs.} The experimental results reveal significant limitations of general-purpose VLMs (GPT4-o, Claude-3.5, Llama-3.2) when applied to astronomical tasks. Their poor performance ($R^2$ scores $< 0.12$ for property estimation) stems from their inherent Euclidean-space constraints, making them ill-suited for modeling universe phenomena in diverse geometric spaces. For instance, their struggle with $\mathbf{sSFR}$ estimation ($R^2$ scores $<$ $0.08$) particularly highlights their inability to capture the hyperbolic nature of star formation dynamics in galaxy evolution.

\textbf{Analysis of Specialized Astronomical Models.} 
Domain-specific approaches demonstrate stronger baseline performance, with AstroCLIP (Fine-tuned) achieving notable results ($R^2 = 0.88$ for $\mathbf{M^*}$ estimation). However, these methods show inconsistent performance across different tasks. While effective in controlled scenarios (e.g., SPR classification: $0.96$ F1), they struggle with complex morphological features requiring geometric understanding (BAR: $0.54$ F1). This performance gap indicates their limited ability to leverage geometric priors and world knowledge, particularly evident in tasks involving spiral arm patterns and bar structures that demand the understanding of galactic rotation dynamics.

\textbf{Performance of GalaxyWalker.}
Our geometry-aware architecture achieves consistent improvements across diverse astronomical tasks through its enhanced spatial modeling capabilities. In property estimation, Galaxy Walker establishes new state-of-the-art benchmarks, notably improving $\mathbf{sSFR}$ estimation ($R^2$ = $0.84$, $+0.15$) through precise modeling of non-Euclidean star formation dynamics. The model exhibits superior performance in complex morphological analysis, particularly in features requiring sophisticated geometric understanding - BAR and SAC classifications both see +0.17 F1 score improvements while maintaining competitive performance in traditional metrics (DEO: $0.97$ F1, SPR: $0.96$ F1). This comprehensive enhancement stems from our geometry prompt's effective modeling of diverse spatial configurations, from spherical planetary orbits to hyperbolic gravitational fields. While showing a marginal decrease in SMH ($-0.07$), the model's substantial improvements in complex geometric features demonstrate its robust capacity for modeling intricate galactic structures. These results, showing $10$-$20$\% average improvement across tasks, validate our geometric integration approach and establish a new paradigm for astronomical vision understanding.

\subsection{Expert Specialization Analysis}
\begin{figure*}[htbp]
    \centering
    \begin{subfigure}[b]{0.36\linewidth}
        \centering
        \includegraphics[width=\linewidth]{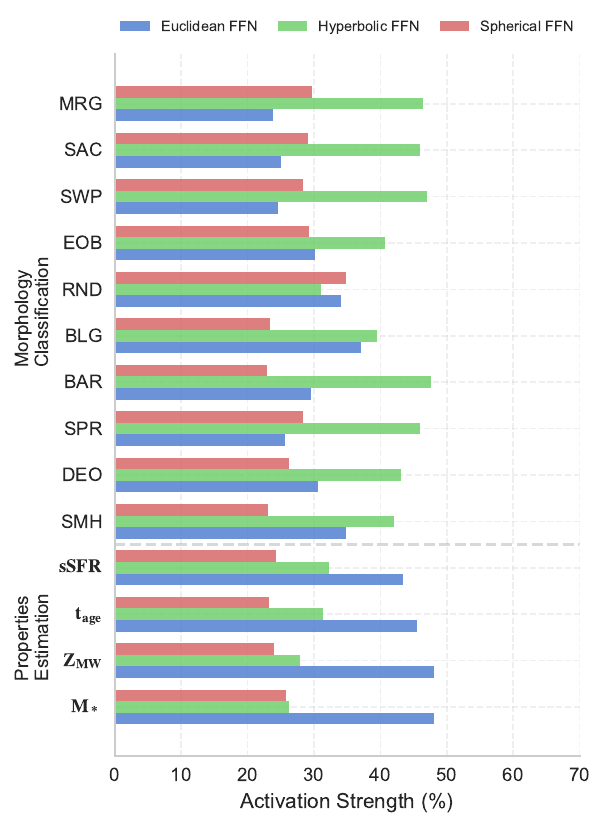}
        \caption{Contribution analysis of different experts. }
        \label{fig:moe_activation}
    \end{subfigure}
    \hfill
    \begin{subfigure}[b]{0.60\linewidth}
        \centering
        \includegraphics[width=\linewidth]{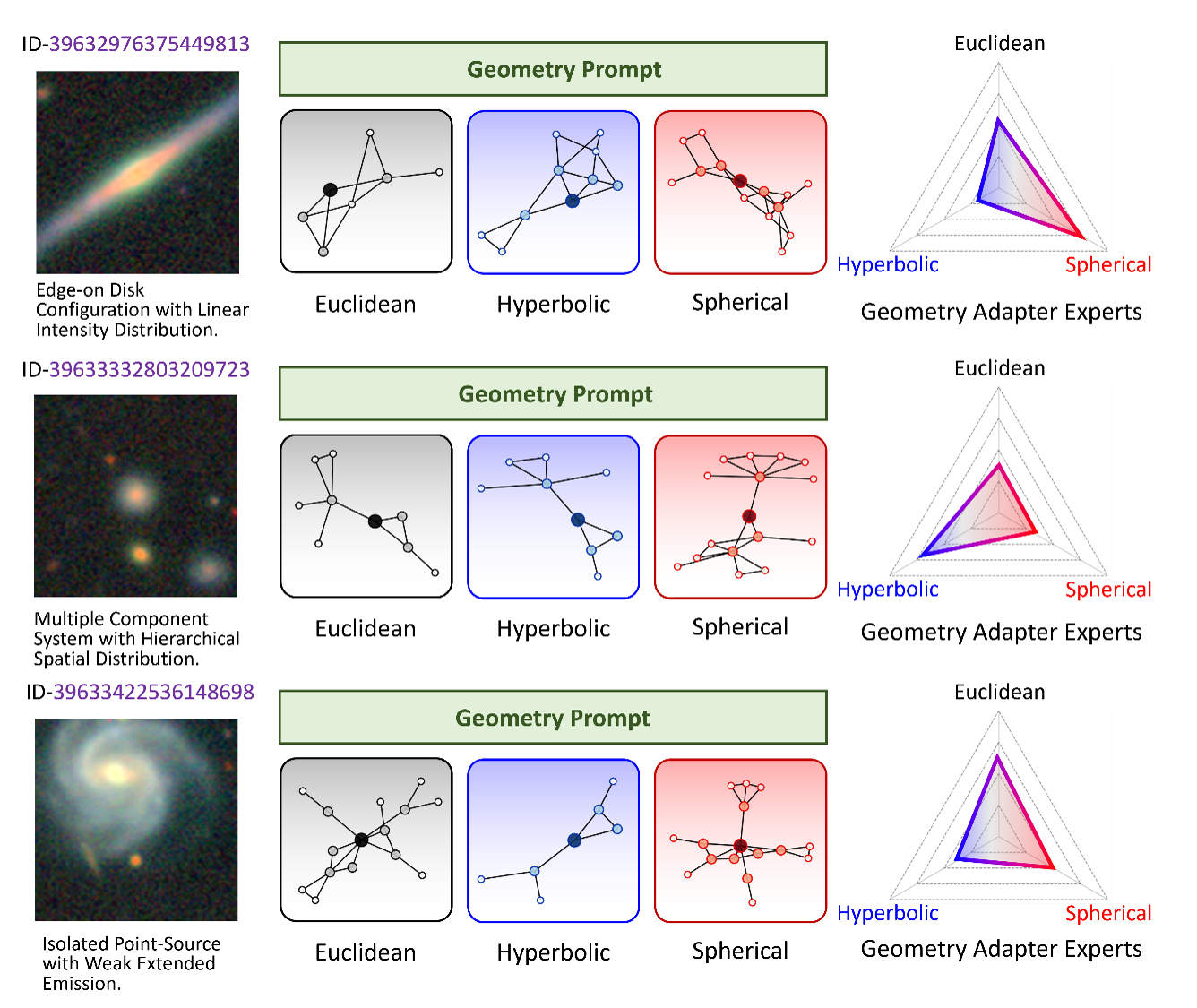}
        \caption{Case study of complex galaxy structures.}
        \label{fig:case_study}
    \end{subfigure}
    \caption{Visualization of geometry-specific expert contributions and case study analysis.}
    \label{fig:expert_analysis}
\end{figure*}

As depicted in Figure \ref{fig:expert_analysis}, to quantitatively analyze the contribution of different geometric experts, we conduct inference on test set and measure each expert's activation strength by averaging normalized weights during the inference process. 

\textbf{Expert Contribution Patterns.} In property estimation tasks, the Euclidean expert demonstrates dominant activation ($>40$\%), particularly in $\mathbf{M^*}$  and $\mathbf{Z_{MW}}$ estimation. This aligns with our assumption that these fundamental galaxy properties primarily rely on direct photometric measurements and local feature extraction, where Euclidean space is most effective. The $\mathbf{t_{age}}$ estimation similarly benefits from Euclidean processing, presumably due to its dependence on spectral energy distribution analysis in conventional space. The morphology classification tasks reveal more diverse geometric preferences. The Hyperbolic expert shows notably higher activation ($>35$\%) in structure-related features like BAR, SPR, and SAC. We hypothesize this is due to the hyperbolic nature of gravitational fields governing these structures - bar formations and spiral arms typically follow logarithmic patterns that are naturally represented in hyperbolic space. The Spherical expert exhibits consistent activation ($20$-$30$\%) across most morphological tasks, presumably capturing the projected 3D spherical nature of galaxies onto our 2D observational plane.

\vspace{-1mm}
\textbf{Case Study Analysis.}
Figure \ref{fig:case_study}  presents three representative cases demonstrating GalaxyWalker's capability in handling diverse galactic topological structures. In the first case, an edge-on disk configuration with linear intensity distribution shows that the Geometry Prompt effectively captures this elongated structure through complementary geometric representations. The triangle plot reveals that the Spherical expert dominates, indicating the model's preference for spherical geometry to model the radial emission patterns and extended structure of the edge-on disk. In the second case, a multiple-component system with hierarchical spatial distribution is analyzed, where our model adapts by increasing the Hyperbolic expert's contribution to effectively represent the complex hierarchical relationships between component galaxies. This preference for hyperbolic geometry enables the model to capture the non-Euclidean nature of gravitational interactions in this system. In the third case of an isolated point source with weak extended emission, the Geometry Adapter shows a more balanced contribution pattern with the Euclidean expert playing a more prominent role compared to previous cases, while still engaging both Spherical and Hyperbolic experts to capture the multifaceted geometric properties. These visualizations quantitatively demonstrate that the proposed geometry-aware framework enables effective encoding of diverse galactic topological patterns while facilitating adaptive geometric feature extraction through dynamically weighted expert contributions.

\subsection{Modality Analysis}

\begin{figure}[htbp]
    \centering
    \begin{subfigure}[b]{0.56\columnwidth}
        \centering
        \includegraphics[width=\columnwidth]{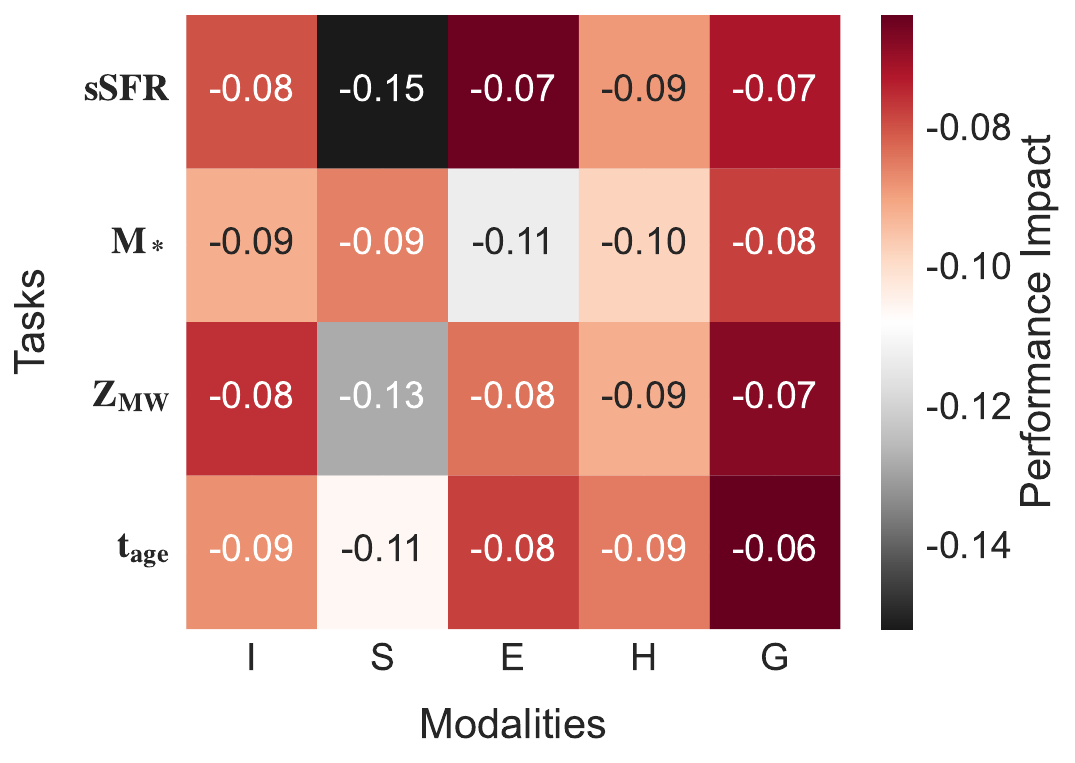}
        \caption{Modality Impact on Different Tasks}
        \label{fig:modal_impact}
    \end{subfigure}
    \hfill
    \begin{subfigure}[b]{0.4\columnwidth}
        \centering
        \includegraphics[width=\columnwidth]{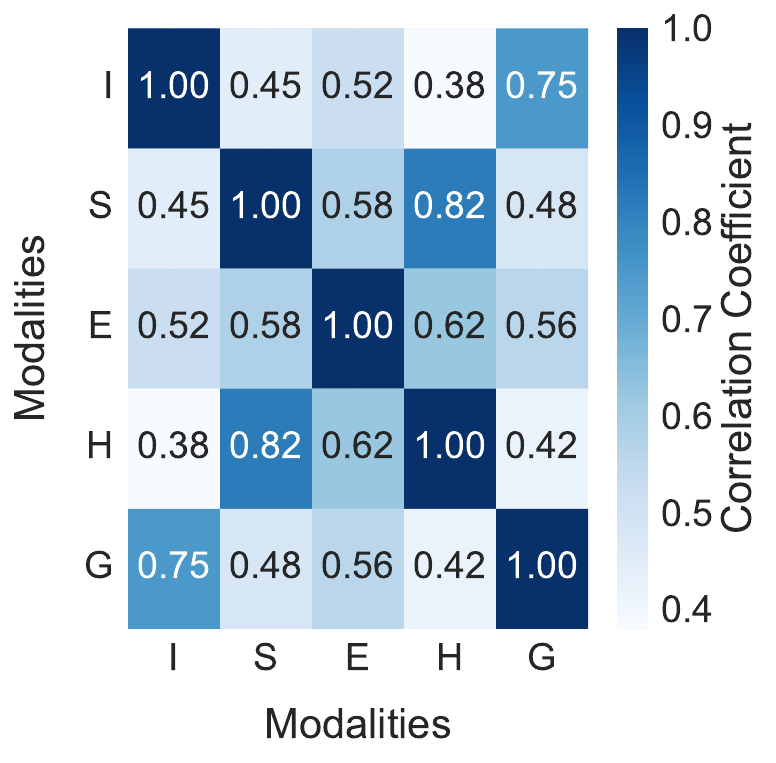}
        \caption{Modality Correlation}
        \label{fig:modal_correlation}
    \end{subfigure}
    \caption{Analysis of modality contributions: (a) Performance impact when removing each modality; (b) Cross-modal correlation analysis.}
    \label{fig:modality_analysis}
\end{figure}

\begin{figure*}[htbp]
    \centering
    \includegraphics[width=0.95\linewidth]{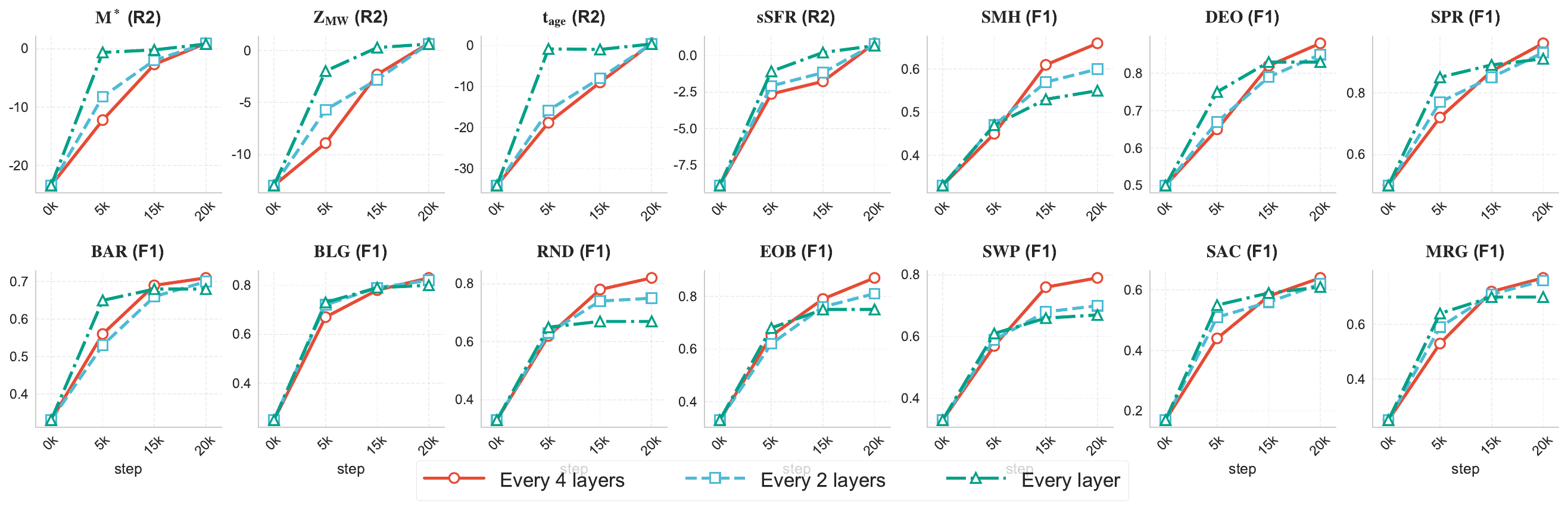}
    \caption{\textbf{Training Dynamics Analysis of Different Geometry Adapter Integration Strategies.} Performance evolution during training ($0$k-$20$k steps) for different adapter integration densities in Qwen2-VL-2B, comparing sparse (every $4$ layer), medium (every $2$ layer), and dense (every layer) integration patterns. The plots show $R^2$ scores for physical property estimation ($\mathbf{M^*}$, $\mathbf{Z_{MW}}$, $\mathbf{t_{age}}$, $\mathbf{sSFR}$) and F1 scores for morphological classification tasks, revealing distinct convergence characteristics across different astronomical tasks.}
    \label{fig:moe_layer}
\end{figure*}

To comprehensively understand the contribution of different modalities in galaxy understanding, we conduct two complementary analyses in Figure \ref{fig:modality_analysis}: an ablation study measuring performance degradation when removing individual modalities, and a correlation analysis examining the relationships between different modality pairs. Our model incorporates five key modalities besides the text modal: Image (I), Spectrum (S), and three geometric representations - Euclidean Graph (E), Hyperbolic Graph (H), and Sphere Graph (G), with the latter three jointly forming our Geometry Prompt Input.

\textbf{Impact of different modality.} As shown in Figure \ref{fig:modal_impact}, removing different modalities impacts model performance to varying degrees across tasks. The spectrum modality demonstrates the strongest impact on \textbf{sSFR} ($-0.15$) and $\mathbf{ Z_{MW}}$ ($-0.13$), reflecting its crucial role in determining both star formation rates and metallicity through spectral line analysis. The Euclidean geometry representation shows moderate influence on $\mathbf{M^*} $ estimation ($-0.11$), while hyperbolic and spherical geometries exhibit relatively consistent impact across all physical parameters ($-0.06$ to $-0.09$). 

\textbf{Cross-modal Correlation.} We use the attention average attention scores between different modal tokens to construct the correlation matrix (Figure \ref{fig:modal_correlation}), which reveals distinctive patterns in modality interactions. Most notably, the high correlation ($0.82$) between spectrum and hyperbolic graph representations suggests that spectral features naturally align with hyperbolic space, likely due to the exponential nature of stellar population distributions. Meanwhile, the strong coupling ($0.75$) between image and sphere graph features indicates that spherical geometric understanding significantly enhances visual feature interpretation, particularly for projecting 3D galactic structures onto 2D observations. The moderate correlation ($0.58$) between the Euclidean graph and spectrum features provides insight into why conventional VLMs can partially succeed in basic spectral analysis tasks, as traditional Euclidean space adequately captures certain spectral characteristics.

\subsection{Geometry Adapter Integration Strategies}

To investigate the optimal integration strategy of geometry-aware capabilities into vision-language models, we conducted experiments with varying densities of MoE-based Geometry Adapter integration in the 28-layer Qwen2-VL-2B architecture~\cite{Qwen}. Figure \ref{fig:moe_layer} reveals distinct patterns in learning dynamics and final performance across different integration frequencies.

Comparative analysis of integration strategies reveals their distinct advantages: sparse integration (every $4$ layer, which is used to achieve results in Table \ref{tab:model_comparison}) achieves competitive performance in morphological classification (DEO: $0.85$, RND: $0.8$) with minimal computational overhead; medium-density integration (every $2$ layer) demonstrates optimal stability across tasks and particularly excels in complex structural feature detection (BAR: $0.7$, BLG: $0.8$); while dense integration (every layer) enables rapid initial convergence but shows diminishing returns in later training stages, suggesting that the additional computational cost may not justify the marginal performance gains.

The training dynamics reveal a clear evolution in the relative advantages of each integration strategy. During early training ($0$-$5$k steps), dense integration provides rapid performance improvements, particularly in property estimation tasks. However, as training progresses ($5$k-$15$k steps), the performance gap between different integration densities diminishes, with all approaches converging to similar performance levels by the final training phase ($15$k-$20$k steps).

Our findings indicate that the optimal integration strategy depends primarily on specific deployment constraints and task priorities. For resource-efficient implementations, sparse integration offers a compelling balance of performance and computational cost. Medium integration provides the most robust general-purpose solution, while dense integration may be warranted in scenarios requiring rapid model adaptation. These results demonstrate that effective geometry-aware vision-language modeling can be achieved through strategic adapter placement, challenging the assumption that comprehensive layer modification is necessary for optimal performance.

\section{Conclusion}
In this paper, we introduced Galaxy Walker, a novel geometry-aware vision-language model that effectively bridges the gap between traditional VLMs and complex astronomical phenomena through the integration of multi-geometric random walks and specialized adapters. By leveraging complementary geometric spaces - Euclidean for local feature extraction, hyperbolic for hierarchical structures, and spherical for celestial projections - our model achieves a more comprehensive understanding of galactic structures. Our approach demonstrates significant performance improvements over existing methods ($10$-$20$\% across tasks), particularly excelling in geometry-intensive challenges like bar structure detection ($+0.17$ F1) and specific star formation rate estimation ($+0.15$ $R^2$). Through comprehensive analysis, we validate our geometric integration strategy and reveal the specialized contributions of different geometric spaces in galaxy understanding, showing how each space contributes distinctively to different aspects of astronomical feature comprehension.

Despite these advances, our work has several limitations. Currently, Galaxy Walker is implemented on relatively small-scale VLMs and employs a limited number of geometric experts. Future work could explore scaling to larger VLMs ($50$B-$100$B parameters) and training domain-specific MoE architectures with expanded expert capacity when sufficient astronomical data becomes available. 

\newpage
\section*{Acknowledgement}
This work was supported by the National Science and Technology Major Project(No.2022ZD0117800), and Young Elite Scientists Sponsorship Program by CAST(No.2023QNRC001). This work was also sponsored by CAAI-Huawei MindSpore Open Fund (CAAIXSJLJJ2023MindSpore12). Yisen Gao is supported by Beijing Natural Science Foundation(QY24129). This work was also developed on Openl Community. Thanks for the computing infrastructure provided by Beijing Advanced Innovation Center for Big Data and Brain Computing. Special thanks for the constructive suggestions from Ming Lu, Shanghang Zhang, Hao Wang, and Jingyang He from Peking University.

{
    \small
    \bibliographystyle{ieeenat_fullname}
    \bibliography{ref}
}

\clearpage
\section*{Appendix}

\section*{A. Training Details}

 The training process consists of two main stages: geometric prompt learning and adapter fine-tuning.

 \subsection*{A.1 Geometry Prompt Learning}

For geometry prompt learning (Stage I), we first construct multi-geometric graph structures using k-nearest neighbors (k=10) in three complementary spaces: Euclidean, Hyperbolic (k=-1), and Spherical (k=1). The input features of dimension 1024 are processed through geometric encoders with a hidden dimension of 512, eventually producing 256-dimensional geometric embeddings for each space. During training, we employ the AdamW optimizer with an initial learning rate of 1e-3 and weight decay of 1e-5. A step-wise learning rate scheduler is applied with step size 100 and decay factor 0.5. To ensure efficient training, we use a batch size of 1024 and train for 100 epochs.

 \subsection*{A.2 Adapter Fine-tuning}
 
For adapter fine-tuning (Stage II), we build upon Qwen2-VL-2B as our base model. The geometry adapter is integrated every four transformer layers, consisting of three types of feed-forward networks: a standard Euclidean FFN inherited from the pre-trained weights, a Hyperbolic FFN based on the Lorentz model with learnable scale, and a Spherical FFN utilizing von Mises-Fisher distribution with learnable concentration. These experts are orchestrated through a gating mechanism with temperature $\tau$ = 0.1. We utilize LoRA for efficient fine-tuning, with rank 16, alpha 32, and dropout 0.1. The training process employs mixed-precision (bf16) and runs with a batch size of 16 per GPU for 3 epochs. We use the AdamW optimizer with a learning rate of 2e-5 and weight decay of 0.01, incorporating 500 warmup steps. The maximum sequence length is set to 512 tokens.

 \subsection*{A.3 Data Pre-processing}
 
For data processing, images are resized to 144×144 pixels and normalized to [0,1] range. Spectral data is encoded into 1024-dimensional features and L2 normalized. These features are then projected into different geometric spaces through manifold-specific mappings, resulting in 256-dimensional representations in each space.

\section*{B. Riemannina Geometry}
A smooth manifold $M$ is referred to as a Riemannian manifold when it possesses a Riemannian metric $g$. Curvature $c$ is an important measure of the degree of geodesic bending.
For each point $x \in M$, there exists a tangent space $T_xM \subseteq \mathbb{R}^d$ that surrounds $x$, where the metric $g$ is applied to determine the manifold's shape. The relationship between the tangent space and the manifold is established through the use of exponential and logarithmic maps. In particular, the exponential map at point $x$, represented as $\exp_x^c(\cdot): T_xM \to M$, transforms points from the tangent space into the manifold, while the logarithmic map function is the inverse function of exponential map $\log_x^c(\cdot)=(\exp_x^c(\cdot))^{-1}$.

In this paper, we use three geometric spaces of different curvature to form a Riemannian expert: Euclidean space ($c=0$), hyperbolic space ($c<0$), and spherical space ($c>0$).

\subsection*{B.1 Euclidean space}

Euclidean space is based on Euclidean coordinates. Since the curvature is zero, the geodesic remains parallel. Euclidean space can be used to describe a flat universe very well. Each galaxy is influenced by its neighbors, capturing the local structure of galaxies in the universe. The exponential mapping of Euclidean Spaces is defined as:
\begin{equation}\exp_{x_p}^c(x)=\mathbf{x}_{p}+\mathbf{x}.\end{equation}

\subsection*{B.2 Hyperbolic space}

A hyperbolic space is defined as $\mathbb{H}_c^{d}{=\{\mathbf{x}_{p}\in\mathbb{R}^{d+1}:\langle\mathbf{x}_{p},\mathbf{x}_{p}\rangle_{\mathcal{L}}=1/c\}},$ where $d$ represents the dimension and the inner product is defined as $\langle\mathbf{x},\mathbf{y}\rangle_{\mathcal{L}}=-x_{1}y_{1}+\sum_{j=2}x_{j}y_{j}$). In a hyperbolic space, The geodesic distance between the two points is:

\begin{equation}
d(x,y)=\frac1{\sqrt{-c}}\operatorname{arccosh}\left(c*\langle\mathbf{x},\mathbf{y}\rangle_{\mathcal{L}}\right).
\end{equation}

Since the curvature is negative, the geodesic will diverge. This helps to describe the evolution of galaxies in a universe, and thus reflects the internal hierarchy of galaxies. Its exponential map is defined as:
\begin{equation}
exp_{x_p}^c(x)=\cosh\left(\sqrt{-c}||\mathbf{x}||\right)\mathbf{x}_p+\sinh\left(\sqrt{-c}||\mathbf{x}||\right)\frac{\mathbf{x}}{\sqrt{-c}||\mathbf{x}||}.
\end{equation}

\subsection*{B.3 Sphere space}
Spherical space is defined as $\mathbb{S}_{c}^{d} = \{\mathbf{x}_{p} \in \mathbb{R}^{d+1} : \langle\mathbf{x}_{p},\mathbf{x}_{p}\rangle_{\mathbb{S}} = 1/c\}$, where the inner product is the standard Euclidean inner product 
$\langle\mathbf{x},\mathbf{y}\rangle_{\mathbb{S}}=\sum_{j=1}^{d+1}x_{j}y_{j}$.The geodesic distance between the two points is:

\begin{equation}d(x,y)=\frac{1}{\sqrt{c}}\arccos\left(c\langle\mathbf{x},\mathbf{y}\rangle_{\mathbb{S}}\right).\end{equation}

Geodesics in spherical space are convergent. Therefore, it can reflect the global information of the galaxy. Capture the overall star map content. Its exponential map is defined as

\begin{equation}
exp_{x_p}^c(x)=\cosh\left(\sqrt{c}||\mathbf{x}||\right)\mathbf{x}_p+\sinh\left(\sqrt{c}||\mathbf{x}||\right)\frac{\mathbf{x}}{\sqrt{-c}||\mathbf{x}||}.
\end{equation}

\section*{C. Special Tokens and Templates}

\subsection*{C.1 Special Modality Tokens}
\label{sec:special_token}
To effectively integrate multi-modal features into the input sequence, we carefully selected special tokens with relatively low frequency in the pre-trained vocabulary to represent different modalities. Specifically, we use token ``\~{A}'' for spectral features, ``$\wp$`` for Euclidean geometric structure, ``\o'' for spherical geometric structure, and ``\ae'' for hyperbolic geometric structure. During forward propagation, these tokens' embeddings are dynamically replaced with their corresponding modal embeddings after geometry-specific projection and normalization. This design allows the model to seamlessly incorporate multi-geometric and spectral information while maintaining the pre-trained model's linguistic capabilities.

\subsection*{C.2 Task-specific Templates}
\label{sec:template}



In this section, we present the task-specific templates for two main categories of tasks: galaxy property estimation and galaxy morphology classification. For each task, we utilize special modality tokens introduced in Section~\ref{sec:special_token} to incorporate different modalities. The key modalities include Image, Spectral (where available), and Geometry information. To better leverage knowledge from pre-trained models, we customize the descriptions of special modality tokens according to each task's characteristics, enabling Galaxy Walker to align multi-modal representations better.

\begin{table*}[t]
\centering
\caption{Options for Galaxy Morphology Classification Tasks}
\label{tab:morphology_options}
\begin{tabular}{ll}
\toprule
\textbf{Task} & \textbf{Options} \\
\midrule
Smooth & (a) Smooth (b) Featured or Disk (c) Artifact \\
Disk-Edge-On & (a) Yes (b) No \\
Spiral-Arms & (a) Yes (b) No \\
Bar & (a) Strong Bar (b) Weak Bar (c) No Bar \\
Bulge-Size & (a) Dominant (b) Large (c) Moderate (d) Small (e) None \\
How-Rounded & (a) Round (b) In-Between (c) Cigar-Shaped \\
Edge-On-Bulge & (a) Boxy (b) None (c) Rounded \\
Spiral-Winding & (a) Tight (b) Medium (c) Loose \\
Spiral-Arm-Count & (a) 1 (b) 2 (c) 3 (d) 4 (e) More than 4 (f) Can't Tell \\
Merging & (a) None (b) Minor Disturbance (c) Major Disturbance (d) Merger \\
\bottomrule
\end{tabular}
\end{table*}

\subsubsection*{C.2.1 Galaxy Property Estimation}

Galaxy property estimation encompasses four regression tasks:
\begin{itemize}
    \item \textbf{Stellar Mass ($\mathbf{M^*}$) Prediction}: For numerical regression of the total mass of stars in a galaxy.
    \item \textbf{Mass-Weighted Stellar Metallicity ($\mathbf{Z_{MW}}$) Prediction}: For estimating the abundance of heavy elements in stars.
    \item \textbf{Mass-Weighted Galaxy Age ($\mathbf{t_{age}}$) Prediction}: For determining the mass-weighted average age of stars (in Gyr).
    \item \textbf{Specific Star-Formation Rate (sSFR) Prediction}: For calculating the rate of star formation per unit stellar mass.
\end{itemize}

For these prediction tasks, we employ a numerical head for regression. In the templates, we use the ``num`` token to represent all numerical values as the model's target response.

\begin{figure*}[t]
\begin{AIbox}{\DV: Stellar Mass Estimation}
{\bf User:}\
{\scriptsize
Stellar mass refers to the total mass of all the stars in a galaxy. It is a critical parameter for understanding galaxy formation and evolution and can be analyzed through multiple perspectives. Specifically, the \textbf{[Image token]} utilizes celestial image data to assess morphology and luminosity, which helps in the initial estimation of stellar mass. The \textbf{[Spectral token]} analyzes stellar spectral characteristics, such as absorption line width and radiation intensity, to directly infer mass parameters. The \textbf{[Euclidean token]} provides the object's position in flat space, aiding in the mass calculation by considering distance measurements. The \textbf{[Hyperbolic token]} describes the geometrical properties in negatively curved space, modeling more complex cosmic structures and helping to understand the distribution of massive stars in a negatively curved universe. The \textbf{[Sphere token]} uses spherical geometry in positively curved space to evaluate an object's position in the spherical coordinate system, leading to a more accurate mass estimation.
}\\
{\bf \textbf{Assistant}:}\
{\scriptsize
NUM
}
\end{AIbox}
\caption{Prompt template for stellar mass estimation.}
\label{fig:prompt_template_stellar_mass}
\end{figure*}

\begin{figure*}[t]
\begin{AIbox}{Galaxy Walker: Mass-Weighted Stellar Metallicity Estimation}
{\bf User:}\
{\scriptsize
Mass-weighted stellar metallicity measures the abundance of elements heavier than hydrogen and helium in a galaxy's stars, weighted by their mass. This aids in understanding the galaxy's chemical evolution and can be analyzed through multiple perspectives. Specifically, the \textbf{[Image token]} helps observe the color and brightness variations of celestial objects, providing initial metallicity estimates. The \textbf{[Spectral token]} offers a detailed analysis of spectral lines, such as the strength and shift of metal lines, to directly infer the mass-weighted metallicity. The \textbf{[Euclidean token]} provides precise coordinates in flat space, aiding in the calculation of metallicity distribution within stars by using distance information. The \textbf{[Hyperbolic token]} describes the geometrical properties in negatively curved space, modeling complex star cluster structures and giving geometrical background support for metallicity distribution. The \textbf{[Sphere token]} employs spherical geometry in positively curved space to understand the distribution of celestial objects within the spherical coordinate system, leading to comprehensive metallicity estimation.
}\
\\
{\bf Assistant:}\
{\scriptsize
NUM
}
\end{AIbox}
\caption{Prompt template for mass-weighted stellar metallicity estimation.}
\label{fig:prompt_template_metallicity}
\end{figure*}

\begin{figure*}[t]
\begin{AIbox}{Galaxy Walker: Mass-Weighted Galaxy Age Estimation}
{\bf User:}\
{\scriptsize
Mass-weighted galaxy age refers to the average age of stars within a galaxy, weighted by their mass, providing insights into the galaxy's formation history. This can be analyzed through multiple perspectives. Specifically, the \textbf{[Image token]} assesses morphology and color via celestial images to estimate the age distribution of stellar populations in the galaxy. The \textbf{[Spectral token]} primarily uses spectral analysis, such as examining the spectral energy distribution and absorption line changes, to determine the overall age of the galaxy. The \textbf{[Euclidean token]} provides the galaxy's coordinates in flat space, assisting in refining age estimation based on distance and position. The \textbf{[Hyperbolic token]} describes complex geometrical backgrounds in negatively curved space, aiding in the detailed understanding of mass-weighted age composition. The \textbf{[Sphere token]} utilizes positively curved space in spherical geometry to assist in distribution analysis and age estimation of different regions within the galaxy.
}\\
{\bf Assistant:}\
{\scriptsize
NUM
}
\end{AIbox}
\caption{Prompt template for mass-weighted galaxy age estimation.}
\label{fig:prompt_template_age}
\end{figure*}

\begin{figure*}[t]
\begin{AIbox}{Galaxy Walker: Specific Star-Formation Rate Estimation}
{\bf User:}\
{\scriptsize
The specific star-formation rate (sSFR) is the rate of star formation per unit stellar mass in a galaxy, indicating how actively the galaxy is forming stars relative to its existing stellar mass. It can be analyzed through multiple perspectives. Specifically, the \textbf{[Image token]} helps analyze star-forming regions, morphology, and density variations via celestial images for initial estimation of the star-formation rate. The \textbf{[Spectral token]} provides detailed spectral analysis, especially the intensity and distribution of emission lines, to measure the current star-formation rate. The \textbf{[Euclidean token]} offers precise positioning in flat space, aiding in inferring the star-formation rate based on distance and velocity information. The \textbf{[Hyperbolic token]} describes geometrical properties in negatively curved space, modeling complex cosmic environments and star cluster structures for supporting star-formation rate estimation. The \textbf{[Sphere token]} utilizes positively curved space in spherical geometry to understand the distribution of star formation within the spherical coordinate system, assisting in specific rate determination.
}\\
{\bf Assistant:}\
{\scriptsize
NUM
}
\end{AIbox}
\caption{Prompt template for specific star-formation rate estimation.}
\label{fig:prompt_template_ssfr}
\end{figure*}

\subsubsection*{C.2.2 Galaxy Morphology Classification}

Galaxy morphology classification includes ten distinct classification tasks:
\begin{enumerate}
    \item Smooth (SMH)
    \item Disk-Edge-On (DEO)
    \item Spiral-Arms (SPR)
    \item Bar (BAR)
    \item Bulge-Size (BLG)
    \item How-Rounded (RND)
    \item Edge-On-Bulge (EOB)
    \item Spiral-Winding (SWP)
    \item Spiral-Arm-Count (SAC)
    \item Merging (MRG)
\end{enumerate}

For classification tasks, we structure the templates as multiple-choice questions, with the model required to select from options labeled (a), (b), (c), etc. The specific options for each classification task are presented in Table~\ref{tab:morphology_options}.

\begin{figure*}[t]
\begin{AIbox}{\DV: Galaxy Smoothness Classification}
{\bf User:}\
{\scriptsize
The morphological class of a galaxy can be analyzed using multiple tokens: the morphological classification of galaxies, such as spiral, elliptical, or irregular, can be directly observed by analyzing their images. Specifically, the \textbf{[Image token]} utilizes celestial images to assess the galaxy's overall shape and structural features, helping to classify it as smooth, featured, or an artifact. The \textbf{[Euclidean token]} offers the galaxy's precise coordinates in flat space, allowing for spatial analysis and comparison with known morphological classes. The \textbf{[Hyperbolic token]} provides insights into negative curvature space, aiding in the understanding of complex structures that might influence the galaxy's morphology. The \textbf{[Sphere token]} uses spherical geometry to interpret the galaxy's appearance in positively curved space, helping to refine its classification.\textbf{ Please choose from these options:(a) Smooth  (b) Featured or Disk  (c) Artifact.}
}\\
{\bf Assistant:}\\
{\scriptsize
[The choice of true label]
}
\end{AIbox}
\caption{Prompt template for galaxy smoothness classification.}
\label{fig:prompt_template_smooth}
\end{figure*}

\begin{figure*}[t]
\begin{AIbox}{\DV: Disk-Edge-On Classification}
{\bf Prompt:}\
{\scriptsize
Determining if a galaxy is disk-edge-on can be analyzed using multiple tokens: edge-on disk galaxies are characterized by their flat, edge-like appearance when observed. This can be directly identified from images. Specifically, the \textbf{[Image token]} offers visual information on the galaxy's edge-on appearance, which is indicative of a disk-edge-on orientation. The \textbf{[Euclidean token]} gives the galaxy's precise coordinates in flat space, assisting in spatial orientation analysis. The \textbf{[Hyperbolic token]} models the galaxy's structure in a negatively curved space, helping to understand any distortions that confirm its disk-edge-on nature. The \textbf{[Sphere token]} uses spherical geometry to analyze the galaxy's orientation in positively curved space. \textbf{Please choose from these options: (a) Yes, it is a disk-edge-on galaxy  (b) No, it is not a disk-edge-on galaxy.}
}\\
{\bf Assistant:}
{\scriptsize
[The choice of true label]
}
\end{AIbox}
\caption{Prompt template for disk-edge-on classification.}
\label{fig:prompt_template_disk_edge_on}
\end{figure*}

\begin{figure*}[t]
\begin{AIbox}{\DV: Spiral Arms Classification}
{\bf Prompt:}\
{\scriptsize
Determining if a galaxy has spiral arms can be analyzed using multiple tokens: spiral-arm galaxies typically exhibit distinct spiral patterns in images. Specifically, the \textbf{[Image token]} provides visual information to identify the presence and patterns of spiral arms. The \textbf{[Euclidean token]} provides the galaxy's coordinates in flat space, aiding in spatial relationship analysis of spiral structures. The \textbf{[Hyperbolic token]} models the galaxy in a negatively curved space, providing geometric context for the spiral arms' formation. The \textbf{[Sphere token]} uses spherical geometry to interpret the distribution and winding of spiral arms in positively curved space. \textbf{ Please choose from these options:(a) Yes, it is a spiral-arms galaxy  (b) No, it is not a spiral-arms galaxy.}
}\\
{\bf Assistant:}\
{\scriptsize
[The choice of true label]
}
\end{AIbox}
\caption{Prompt template for spiral arms classification.}
\label{fig:prompt_template_spiral_arms}
\end{figure*}

\begin{figure*}[t]
\begin{AIbox}{\DV: Bar Type Classification}
{\bf Prompt:}\
{\scriptsize
Determining the type of bar in a galaxy can be analyzed using multiple tokens: bar structures in galaxies can be directly observed through images, revealing their length and strength. Specifically, the \textbf{[Image token]} offers visual information to observe and classify the bar's strength in the galaxy. The \textbf{[Euclidean token]} gives the galaxy's position in flat space, assisting in the spatial analysis of the bar. The \textbf{[Hyperbolic token]} helps model the galaxy's structure in negatively curved space, which aids in understanding the bar type. The \textbf{[Sphere token]} uses spherical geometry to analyze the distribution of stellar masses in the bar, refining its classification.\textbf{ Please choose from these options: (a) Strong Bar  (b) Weak Bar  (c) No Bar.}
}\\
{\bf Assistant:}\
{\scriptsize
[The choice of true label]
}
\end{AIbox}
\caption{Prompt template for bar type classification.}
\label{fig:prompt_template_bar}
\end{figure*}

\begin{figure*}[t]
\begin{AIbox}{\DV: Bulge Size Classification}
{\bf Prompt:}\
{\scriptsize
Determining the bulge size of a galaxy can be analyzed using multiple tokens: the size of a galaxy's bulge can be observed in images by its prominence. Specifically, the \textbf{[Image token]} provides visual information to assess the bulge's prominence in the galaxy. The \textbf{[Euclidean token]} offers the galaxy's coordinates in flat space, assisting in spatial analysis of the bulge's physical size. The \textbf{[Hyperbolic token]} models the galaxy in negatively curved space, which helps in understanding the bulge size in a broader context. The \textbf{[Sphere token]} uses spherical geometry to analyze the distribution and density of stars within the bulge.\textbf{ Please choose from these options:(a) Dominant Bulge  (b) Large Bulge  (c) Moderate Bulge  (d) Small Bulge  (e) No Bulge.}
}\\
{\bf Assistant:}\
{\scriptsize
[The choice of true label]
}
\end{AIbox}
\caption{Prompt template for bulge size classification.}
\label{fig:prompt_template_bulge_size}
\end{figure*}

\begin{figure*}[t]
\begin{AIbox}{\DV: Galaxy Roundness Classification}
{\bf Prompt:}\
{\scriptsize
Determining the shape of a galaxy can be analyzed using multiple tokens: the shape of a galaxy can be directly observed by analyzing its images. Specifically, the \textbf{[Image token]} provides visual information to classify the galaxy as round, in-between, or cigar-shaped. The \textbf{[Euclidean token]} gives the galaxy's precise coordinates in flat space, aiding in the geometric analysis of its shape. The \textbf{[Hyperbolic token]} models the galaxy's structure in negatively curved space, providing a complex geometric context for its shape classification. The \textbf{[Sphere token]} uses spherical geometry to analyze the galaxy's three-dimensional shape in positively curved space. \textbf{ Please choose from these options: (a) Round  (b) In-Between  (c) Cigar-Shaped.}
}\\
{\bf Assistant:}\
{\scriptsize
[The choice of true label]
}
\end{AIbox}
\caption{Prompt template for galaxy roundness classification.}
\label{fig:prompt_template_roundness}
\end{figure*}

\begin{figure*}[t]
\begin{AIbox}{\DV: Edge-On Bulge Classification}
{\bf Prompt:}\
{\scriptsize
Determining the type of bulge in an edge-on galaxy can be analyzed using multiple tokens: the type of bulge in an edge-on galaxy can be identified by observing its images, which show whether it is boxy or rounded. Specifically, the \textbf{[Image token]} gives visual information to identify and classify the bulge as boxy, rounded, or absent in an edge-on galaxy. The \textbf{[Euclidean token]} offers the galaxy's coordinates in flat space, assisting in the spatial analysis of the bulge. The \textbf{[Hyperbolic token]} models the galaxy in negatively curved space, helping to understand the bulge type in a broader geometrical context. The \textbf{[Sphere token]} uses spherical geometry to analyze the three-dimensional distribution of stars within the bulge.\textbf{ Please choose from these options:(a) Boxy Bulge  (b) No Bulge  (c) Rounded Bulge.}
}\\
{\bf Assistant:}\
{\scriptsize
[The choice of true label]
}
\end{AIbox}
\caption{Prompt template for edge-on bulge classification.}
\label{fig:prompt_template_edge_on_bulge}
\end{figure*}

\begin{figure*}[t]
\begin{AIbox}{\DV: Spiral Winding Classification}
{\bf Prompt:}\
{\scriptsize
Analyzing how tightly wound the spiral arms of a galaxy are can be done using multiple tokens: the tightness of spiral arms can be directly observed in images, showing the winding patterns clearly. Specifically, the \textbf{[Image token]} provides visual information to determine the tightness of the spiral arms. The \textbf{[Euclidean token]} gives the galaxy's precise coordinates in flat space, aiding in the spatial analysis of spiral arm winding. The \textbf{[Hyperbolic token]} models the galaxy's structure in negatively curved space, helping to understand the geometric properties affecting spiral arm tightness. The \textbf{[Sphere token]} uses spherical geometry to analyze the three-dimensional winding of the spiral arms. \textbf{Please choose from these options:(a) Tight Winding  (b) Medium Winding  (c) Loose Winding.}
}\\
{\bf Assistant:}\
{\scriptsize
[The choice of true label]
}
\end{AIbox}
\caption{Prompt template for spiral winding classification.}
\label{fig:prompt_template_spiral_winding}
\end{figure*}

\begin{figure*}[t]
\begin{AIbox}{\DV: Spiral Arm Count Classification}
{\bf Prompt:}\
{\scriptsize
Determining the number of spiral arms in a galaxy can be analyzed using multiple tokens: the number of spiral arms in a galaxy can be directly counted from images. Specifically, the \textbf{[Image token]} provides visual information to count and identify the number of spiral arms. The \textbf{[Euclidean token]} gives the galaxy's precise coordinates in flat space, aiding in the spatial analysis of the spiral arms. The \textbf{[Hyperbolic token]} models the galaxy's structure in negatively curved space, providing a geometric context for the number of spiral arms. The \textbf{[Sphere token]} uses spherical geometry to analyze the three-dimensional distribution of spiral arms. \textbf{Please choose from these options:(a) 1 Spiral Arm  (b) 2 Spiral Arms  (c) 3 Spiral Arms  (d) 4 Spiral Arms  (e) More than 4 Spiral Arms  (f) Can't Tell.}
}\\
{\bf Assistant:}\
{\scriptsize
[The choice of true label]
}
\end{AIbox}
\caption{Prompt template for spiral arm count classification.}
\label{fig:prompt_template_spiral_arm_count}
\end{figure*}

\begin{figure*}[t]
\begin{AIbox}{\DV: Galaxy Merging State Classification}
{\bf Prompt:}\
{\scriptsize
Determining the merging state of a galaxy can be analyzed using multiple tokens: the merging state of a galaxy can be observed through signs of disturbance or merging in images. Specifically, the \textbf{[Image token]} provides visual information to observe signs of merging or disturbances. The \textbf{[Euclidean token]} offers the galaxy's coordinates in flat space, aiding in assessing merging stages from spatial data. The \textbf{[Hyperbolic token]} models the galaxy in negatively curved space, helping to understand the geometric properties affecting the merging state. The \textbf{[Sphere token]} uses spherical geometry to analyze the three-dimensional interactions of merging galaxies\textbf{. Please choose from these options: (a) No Merging  (b) Minor Disturbance  (c) Major Disturbance  (d) Merger.}
}\\
{\bf Assistant:}
{\scriptsize
[The choice of true label]
}
\end{AIbox}
\caption{Prompt template for galaxy merging state classification.}
\label{fig:prompt_template_merging}
\end{figure*}

\section*{D.Additional Experiments}

\begin{figure*}[htbp]
\centering
\includegraphics[width=0.92\linewidth]{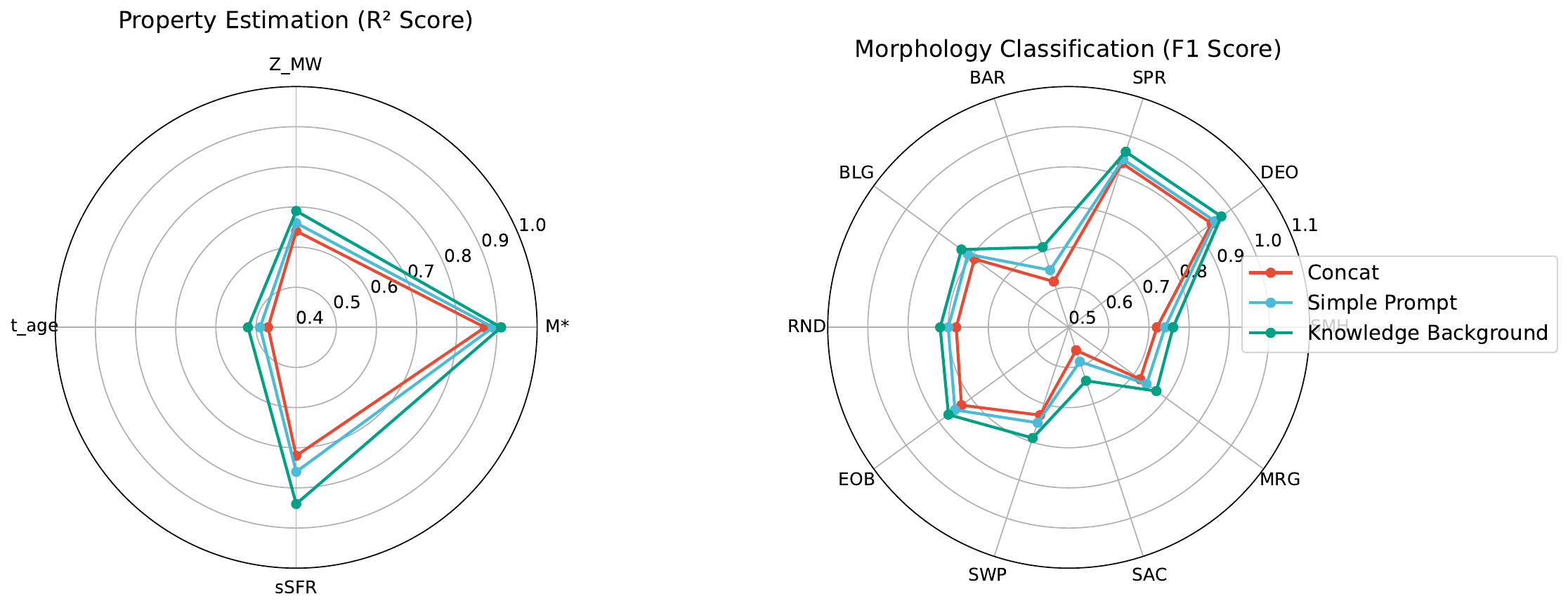}
\caption{\textbf{Performance comparison of different prompt settings.} The radar plots show the performance of three prompt settings across property estimation tasks (left) and morphology classification tasks (right). The Knowledge Background prompts consistently outperform simpler approaches, demonstrating the effectiveness of incorporating domain knowledge into prompts. The improvement is particularly notable in property estimation tasks, where the Knowledge Background setting achieves superior performance in all metrics, with the most significant gains in sSFR estimation. For morphology classification, while the margins are smaller, the Knowledge Background setting still shows consistent advantages, especially in complex features like Bar and Spiral Arm Count classification.}
\label{fig:prompt_comparison}
\end{figure*}

\subsection*{D.1 The Impact of Prompt}
To investigate whether our designed prompts can effectively leverage knowledge from pre-trained models to improve performance, we conduct experiments comparing three different prompt settings:

\begin{itemize}
\item \textbf{Concat}: Directly concatenating modality tokens (Image Token, Spectral Token, Geometry Token) with the question text without any connecting words or explanations.
\item \textbf{Simple Prompt}: Adding basic connecting words to describe what each modality token represents, building upon the Concat setting.
\item \textbf{Prompt with Knowledge Background}: Using our carefully designed templates from Section~\ref{sec:template}, which incorporate detailed explanations of how each modality token contributes to the specific task, combined with relevant domain knowledge.
\end{itemize}

As shown in Figure~\ref{fig:prompt_comparison}, the Prompt with Knowledge Background setting consistently outperforms the other two approaches. In property estimation tasks, this setting achieves notably better results across all metrics, with particular improvements in sSFR prediction. For morphology classification tasks, while all three settings perform competitively, the Knowledge Background prompts still demonstrate advantages, especially in complex features such as Bar and Spiral Arm Count classification. These results suggest that carefully designed prompts incorporating domain knowledge can effectively help the model leverage pre-trained knowledge for better task performance.

\subsection*{D.2 Parameter-Efficient Training Strategy Analysis}

We evaluate three different training strategies to investigate the effectiveness of our parameter-efficient approach:
\begin{itemize}
\item \textbf{Geometry Adapter Only}: Training only the newly added components including projection $\pi_\theta$, Geometry Adapter, and Num Head weights.
\item \textbf{Geometry Adapter + LoRA}: Training the Geometry Adapter components plus LoRA modules in attention and linear layers.
\item \textbf{Full-Parameter Training}: Fine-tuning all model parameters.
\end{itemize}
As shown in Table~\ref{tab:training_strategy}, our Geometry Adapter + LoRA strategy achieves comparable or even superior performance to full-parameter training across most metrics. Notably, it outperforms full-parameter training in property estimation tasks, achieving better $R^2$ scores for all four properties ($\mathbf{M_*}$, $\mathbf{Z_{MW}}$, $\mathbf{t_{age}}$, \textbf{sSFR}). For morphology classification, the performance difference is minimal, with our parameter-efficient approach showing slight advantages in several categories (DEO, SPR, BAR).
The Geometry Adapter Only setting, while using the fewest trainable parameters, still maintains strong performance, suggesting that the geometric adaptation components effectively capture domain-specific features. These results demonstrate that our parameter-efficient strategy can match or exceed the performance of full-parameter training while significantly reducing the number of trainable parameters and computational cost.

\begin{table}[htbp]
\centering
\begin{tabular}{lc}
\toprule
\textbf{Hardware} & \textbf{Inference Time (s)} \\
\midrule
NVIDIA H100 & 0.38 \\
NVIDIA A100 & 1.14 \\
Ascend 910B\textsuperscript{} & 1.52 \\
\bottomrule
\end{tabular}
\caption{\textbf{Inference time comparison across different hardware platforms.} Times are averaged over 100 runs with batch size 1. \textsuperscript{}Ascend 910B results are measured using FP16 precision, while NVIDIA results use BF16 precision.}
\label{tab:inference_time}
\end{table}

\begin{table*}[htbp]
    \centering
    \renewcommand{\arraystretch}{1.0}
    \resizebox{\textwidth}{!}{
        \begin{tabular}{l|cccc|cccccccccc}
            \toprule
            \multirow{2}{*}{\textbf{Training Strategy}} & \multicolumn{4}{c|}{\textbf{Property Estimation ($R^2$ Score)}} & \multicolumn{10}{c}{\textbf{Morphology Classification (F1 Score)}} \\
            \cmidrule(lr){2-5} \cmidrule(lr){6-15}
            & $\mathbf{M_*}$ & $\mathbf{Z_{MW}}$ & $\mathbf{t_{age}}$ & \textbf{sSFR} & \textbf{SMH} & \textbf{DEO} & \textbf{SPR} & \textbf{BAR} & \textbf{BLG} & \textbf{RND} & \textbf{EOB} & \textbf{SWP} & \textbf{SAC} & \textbf{MRG} \\
            \midrule
            Geometry Adapter Only & 0.89 & 0.67 & 0.50 & 0.81 & 0.74 & 0.95 & 0.94 & 0.68 & 0.81 & 0.80 & 0.85 & 0.77 & 0.62 & 0.75 \\
            Geometry Adapter + LoRA & \textbf{0.91} & \textbf{0.69} & \textbf{0.52} & \textbf{0.84} & 0.76 & \textbf{0.97} & \textbf{0.96} & \textbf{0.71} & \textbf{0.83} & \textbf{0.82} & \textbf{0.87} & \textbf{0.79} & \textbf{0.64} & \textbf{0.77} \\
            Full-Parameter Training & 0.90 & 0.68 & 0.51 & 0.82 & \textbf{0.77} & 0.96 & 0.95 & 0.69 & 0.82 & 0.81 & 0.86 & 0.78 & 0.63 & 0.76 \\
            \bottomrule
        \end{tabular}
    }
    \caption{\textbf{Comparison of different training strategies.} Results show that our parameter-efficient approach (Geometry Adapter + LoRA) achieves comparable or even superior performance to full-parameter training while requiring significantly fewer trainable parameters. The best results for each metric are shown in \textbf{bold}.}
    \label{tab:training_strategy}
\end{table*}

\subsection*{D.3 Inference Time Analysis}
To evaluate the practical deployment potential of GalaxyWalker, we conduct inference time benchmarks across different hardware platforms. We measure the average inference time per sample using batch size 1, with BF16 precision on NVIDIA GPUs and FP16 precision on Ascend hardware.

As shown in Table~\ref{tab:infer_time}, our framework maintains practical efficiency through two key design choices: (1) Geometric-aware module allows operating the sparse topologies in geometric hidden spaces directly with O(N) complexity, and adding only 0.08-0.10s latency across hardware platforms. (2) The MoE architecture achieves 87\% sparsity in expert activation through learned gating, limiting its overhead to 0.38-1.52s. The total inference time remains under 1.62s even on mid-tier hardware (Ascend 910B), demonstrating that our multi-geometric modeling adds minimal computational burden while enabling significant performance gains.

\begin{table}[!hbp]
\caption{\textbf{Inference Time of Our Galaxy Walker.}}
\vspace{-3mm}
\renewcommand{\arraystretch}{0.8}
\centering
\label{tab:infer_time}
\begin{tabular}{cccc}
\toprule
\textbf{Hardware} & \textbf{Geo. } & \textbf{MoE.} & \textbf{Total Time(s)} \\ \midrule
NVIDIA H100       & 0.08                     & 0.38                  & 0.46                       \\
NVIDIA A100       & 0.09                     & 1.14                  & 1.23                       \\
Ascend 910B       & 0.10                     & 1.52                  & 1.62                       \\ \bottomrule
\end{tabular}
\end{table}

\subsection*{D.4 More baselines}
Our experiment (Table~\ref{tab:baseline}) reveals: (1) Naive fine-tuning of Qwen2-VL underperforms even the domain-specific AstroCLIP in classification (F1: 0.742 vs 0.767), indicating standard VLM adaptation struggles with astronomical features. (2) Galaxy Walker achieves 10.4\% higher R$^2$ and 5.8\% better F1 than AstroCLIP, highlighting the critical role of geometric modeling in bridging the domain gap for VLMs. This validates our physics-guided architecture against both generalist and specialist baselines. 

\begin{table}[!hbp]
\centering
\caption{\textbf{Compare with naively fine-tuning VLM.}}
\label{tab:baseline}
\vspace{-3mm}
\begin{tabular}{ccc}
\toprule
\textbf{Models}             & \textbf{Estim.(R$^2$)} & \textbf{Classifi.(F1)} \\ \midrule
AsctroCLIP(Baseline)        & 0.670                            & 0.767                       \\
Qwen2-VL(Navie FT)          & 0.693                            & 0.742                       \\
\textbf{GalaxyWalker(Ours)} & \textbf{0.740}                   & \textbf{0.812}              \\ \bottomrule
\end{tabular}
\end{table}

\section*{E. Limitation}

Our method is primarily designed for astronomical data characterized by Riemannian geometric properties, serving as a general-purpose visual model tailored for astronomical tasks. Consequently, it may not be directly applicable to data from other domains that lack distinct geometric properties or cosmic graph structures.


\end{document}